\documentclass[letterpaper, 10 pt, conference]{ieeeconf}
\IEEEoverridecommandlockouts    
\usepackage{graphics}           
\usepackage{times}              
\usepackage{amsmath}            
\usepackage{amssymb}            
\usepackage{graphicx}
\usepackage{algorithm}
\usepackage[noend]{algpseudocode}
\usepackage{booktabs}
\usepackage{color}
\definecolor{instructioncolor}{rgb}{.5,.5,.5}

\usepackage[font=small]{caption}

\def\figref#1{Fig.~\ref{#1}}
\def\tabref#1{Tab.~\ref{#1}}
\def\eqref#1{Eq.~(\ref{#1})}

\makeatletter
\usepackage{xspace}
\DeclareRobustCommand\onedot{\futurelet\@let@token\@onedot}
\def\@onedot{\ifx\@let@token.\else.\null\fi\xspace}

\def\etal{{et al}\onedot}
\makeatother

\def\etalcite#1{\etal~\cite{#1}}

\usepackage{array}
\newcolumntype{L}[1]{>{\raggedright\let\newline\\\arraybackslash\hspace{0pt}}m{#1}}
\newcolumntype{C}[1]{>{\centering\let\newline\\\arraybackslash\hspace{0pt}}m{#1}}
\newcolumntype{R}[1]{>{\raggedleft\let\newline\\\arraybackslash\hspace{0pt}}m{#1}}

\newcommand{\norm}[1]{\lVert#1\lVert}

\renewcommand{\b}[1]{\mbox{\boldmath$#1$}}

\usepackage{subcaption}

\title{\LARGE \bf Robust Double-Encoder Network for RGB-D Panoptic Segmentation}

\author{Matteo Sodano \and Federico Magistri \and Tiziano Guadagnino \and Jens Behley \and Cyrill Stachniss
  \thanks{All authors are with the University of Bonn, Germany. C.~Stachniss is also with the Department of Engineering Science at the University of Oxford, UK, and with the Lamarr Institute for Machine Learning and Artificial Intelligence, Germany.}%
  \thanks{This work has partially been funded
  by the European Union’s Horizon 2020 research and innovation programme under grant agreement No~101017008~(Harmony).
  }%
}

\begin{document}
\maketitle  
\thispagestyle{empty}
\pagestyle{empty}

\begin{abstract}
  %
  Perception is crucial for robots that act in real-world environments, as autonomous systems need to see and understand the world around them to act properly. Panoptic segmentation provides an interpretation of the scene by computing a pixelwise semantic label together with instance IDs. In this paper, we address panoptic segmentation using \mbox{RGB-D} data of indoor scenes. We propose a novel encoder-decoder neural network that processes RGB and depth separately through two encoders. The features of the individual encoders are progressively merged at different resolutions, such that the RGB features are enhanced using complementary depth information. We propose a novel merging approach called ResidualExcite, which reweighs each entry of the feature map 
  according to its importance. With our double-encoder architecture, we are robust to missing cues. In particular, the same model can train and infer on \mbox{RGB-D}, RGB-only, and depth-only input data, without the need to train specialized models. We evaluate our method on publicly available datasets and show that our approach achieves superior results compared to other common approaches for panoptic segmentation.
\end{abstract}

\begin{figure}[t]
  \centering
  \begin{subfigure}{.88\linewidth}
    \centering
    \includegraphics[width=1.1\linewidth]{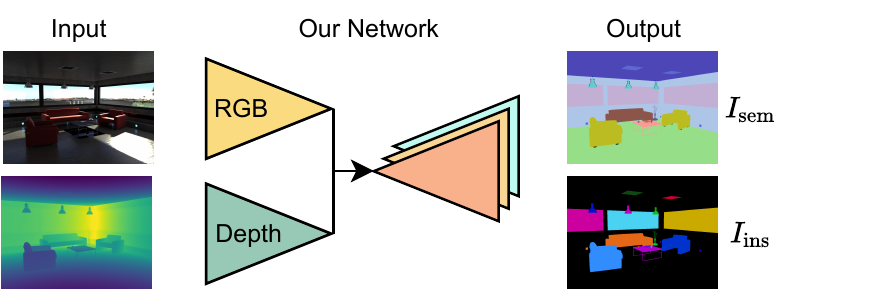}
    \caption{}
    \label{fig:motivation_a}
  \end{subfigure}\\
  \begin{subfigure}{.88\linewidth}
    \centering
    \includegraphics[width=1.1\linewidth]{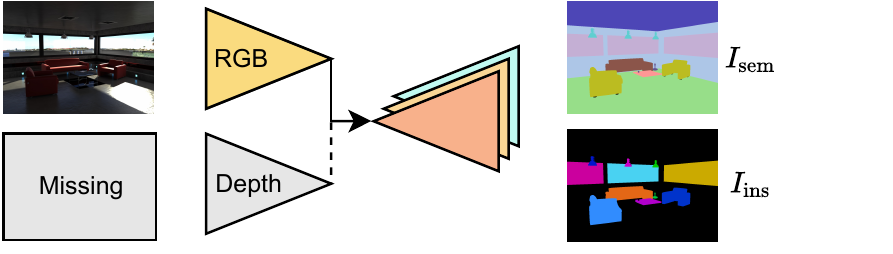}
    \caption{}
    \label{fig:motivation_b}
  \end{subfigure}
  \begin{subfigure}{.88\linewidth}
    \centering
    \includegraphics[width=1.1\linewidth]{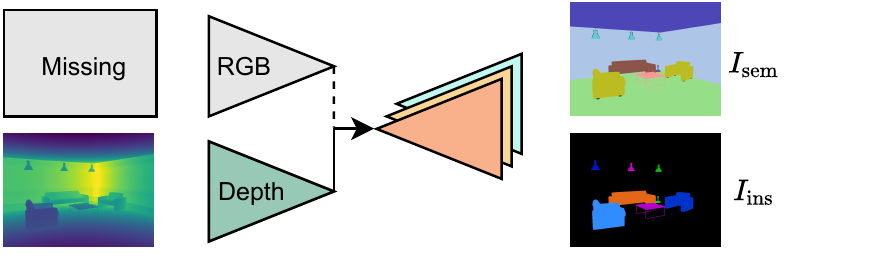}
    \caption{}
    \label{fig:motivation_c}
  \end{subfigure}
  \caption{Our double-encoder network for RGB-D panoptic segmentation is able to provide predictions dealing with full RGB-D images (a), RGB-only (b) or depth-only (c). Dashed 
  lines indicates a detached encoder.}
  \label{fig:motivation}
  \vspace{-2mm}
\end{figure}

\section{Introduction}
\label{sec:intro}

Holistic scene understanding is crucial in several robotics applications.  The ability of recognizing objects and obtaining 
a semantic interpretation of the surrounding environment is one of the key capabilities of truly autonomous systems.
Semantic scene perception and understanding supports several robotics tasks such as 
mapping~\cite{dube2017icra,palazzolo2019iros}, place recognition~\cite{garg2018icra}, and 
manipulation~\cite{schwarz2017ijrr}.
Panoptic segmentation~\cite{kirillov2019cvpr} 
unifies semantic and instance segmentation, and solves both jointly. Its goal is to assign a semantic label 
and an instance ID to each pixel of an image. The content of an image is typically divided into two
sets: \textit{things} and \textit{stuff}. Thing classes are composed of countable objects (such as person, car, 
table), while stuff classes are amorphous regions of space without individual instances (such as sky, street, floor). 

In this paper, we target panoptic segmentation using \mbox{RGB-D} sensors. This data is especially interesting
in indoor environments where the geometric information provided by 
the depth can help dealing with challenging scenarios such as cluttered scenes and dynamic objects.
Additionally, we address the problem of being robust to missing cues, i.e., when either the 
RGB or the depth image is missing. This is a practical issue, as 
robots can be equipped with both, \mbox{RGB-D} and RGB cameras, and sometimes have to operate in poor lighting conditions in which 
RGB data is not reliable. Thus, a single model for handling RGB-D, RGB, and depth data is helpful in practical applications.
We investigate how an encoder-decoder architecture with two encoders for the RGB and depth cues
can provide compelling results in indoor scenes. Previous efforts showed how 
double-encoder architectures are effective in processing \mbox{RGB-D} data~\cite{park2017iccv,seichter2021icra}, 
but they target only semantic segmentation. \\
\begin{figure*}[ht]
  \centering
  \includegraphics[width=0.85\textwidth]{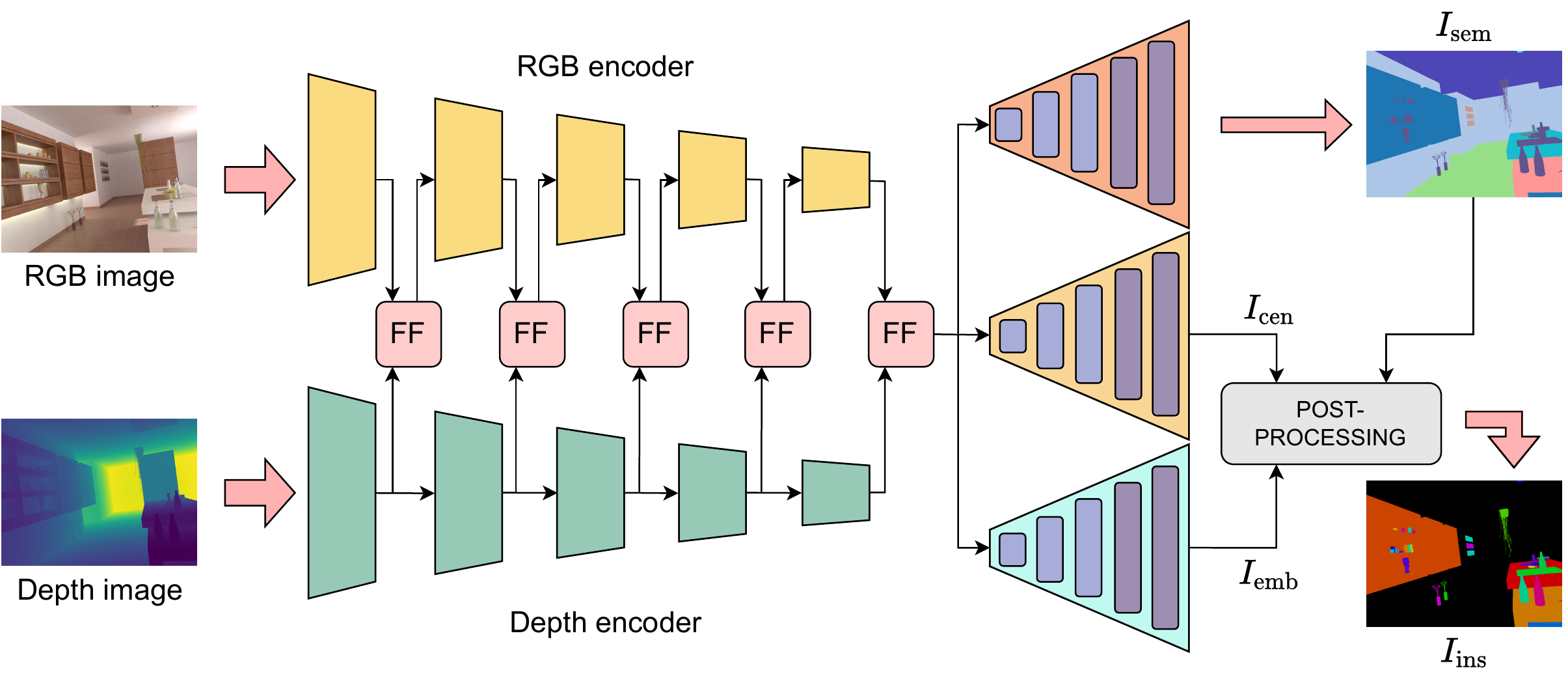}
  \caption{Our double-encoder network for RGB-D panoptic segmentation. RGB and depth images are separately processed, 
  and their features are merged at different output strides by the feature fusion modules (FF).}
  \label{fig:architecture}
\end{figure*}
\indent The main contribution of this paper is a novel approach for \mbox{RGB-D} panoptic segmentation based on a  
double-encoder architecture. 
We propose a novel feature merging strategy, called ResidualExcite, and 
a double-encoder structure robust to missing cues that allows
training and inference with \mbox{RGB-D}, RGB-only, and depth-only data at the same time, without the need to re-train the model
(see \figref{fig:motivation}). 
We show that (i) our fusion mechanism performs better with respect to other state-of-the-art fusion modules, and (ii)
our architecture allows training and inference on RGB-D, RGB-only and depth-only data without the need of a dedicated model for 
each modality.
To back up these claims, we report extensive experiments on the ScanNet~\cite{dai2017cvpr} and 
HyperSim~\cite{roberts2021iccv-haps} datasets. To support reproducibility, our code and dataset splits used in this paper are published at \texttt{https://github.com/PRBonn/PS-res-excite}.




\section{Related Work}
\label{sec:related}

With the advent of deep learning, we witnessed a tremendous progress in the capabilities to provide scene 
interpretation for autonomous robots. 
Kirillov \etalcite{kirillov2019cvpr-ps} define the task of panoptic segmentation as 
the combination between semantic and instance segmentation. The goal of this task is to assign a class label to every pixel
and to additionally segment objects instances. 
Most of the approaches targeting panoptic segmentation on images tackle it top-down, as they 
rely on bounding box-based object proposals~\cite{he2017iccv-mr,kirillov2019cvpr}. Their goal 
is to extract a number of candidate object regions~\cite{girshick2014cvpr,hosang2015pami}, and then evaluate them independently.
These methods are effective but they can lead to overlapping segments in the instance prediction. 
In this work, we follow bottom-up approaches~\cite{cheng2020cvpr-pass,gao2019iccv,roggiolani2022icra},
not relying on bounding boxes but operating directly at a pixel level. 

The works mentioned so far use RGB images. Panoptic segmentation is common also for LiDAR data, both in form of range
images~\cite{milioto2020iros} and point clouds~\cite{gasperini2021ral}. However,
when considering \mbox{RGB-D} data, semantic segmentation~\cite{dai2018eccv-jpfs,qi2017iccv-gnnf} 
and instance segmentation~\cite{engelmann2020cvpr,hou2019cvpr-siso} are common, while panoptic segmentation has 
received less attention so far~\cite{narita2019iros,wu2021cvpr}.
The most common ways of elaborating \mbox{RGB-D} data rely on 3D representations via 
truncated signed distance functions~\cite{hou2019cvpr-siso} or voxel grids~\cite{han2020cvpr-oois}. Few works go in the 
direction of using directly \mbox{RGB-D} images. In our 
approach, we target panoptic segmentation directly on \mbox{RGB-D} frames. \\
\indent Double-encoder architectures are the most successful way for processing 2D representations of \mbox{RGB-D} frames. 
They allow to process RGB and depth cues separately with individual encoders and rely on feature fusion for combining
the outputs of the 
encoders~\cite{park2017iccv-rrmr, seichter2021icra}. 
An alternative to the direct exploitation of RGB and depth, 
proposed by Gupta~\etalcite{gupta2014eccv}, consist in a pre-processing of the depth to encode it with 
three channels for each pixel, 
describing horizontal disparity, height above ground and angle between the pixel's surface normal and the gravity direction. 
The core idea of all these works, however, is that RGB and depth are processed separately and fusion happens 
only at a later point in the network, after the encoding part (late fusion).
Hazirbas~\etalcite{hazirbas2016accv}, however, show that feature merging at different feature resolutions can
enhance performance (early-mid fusion). In contrast, we propose to use multi-resolution merging at every 
downsampling step of the encoder. \\
\indent Different merging strategies for features of data streams are available.
Summation~\cite{hazirbas2016accv} and 
concatenation~\cite{loghmani2019ral} are the earliest strategies, which 
have the limit of considering all features without weighing them according to their 
effective usefulness. Newest efforts go in the direction of Squeeze-and-Excitation modules~\cite{seichter2021icra} 
and gated fusion~\cite{xu2021iccv-rada}, which are two different channel-attention mechanisms that aim 
to increase the focus on features that are more relevant. Other works exploit correlations between modalities to 
recalibrate feature maps based on the most informative features~\cite{sirohi2021tro, valada2019ijcv}. 
In our work, we build on top of channel-attention mechanisms.
We propose a new merging mechanism called ResidualExcite, 
inspired by Squeeze-and-Excitation and residual networks~\cite{he2016cvpr}, that aims to measure the importance
of features at a more fine-grained scale. 

Additionally, we leverage the double-encoder structure to have a single model capable of training and inferring 
on different modalities (RGB-D, RGB-only, depth-only). Multi-modal models have been investigated in the past, 
but mostly exploiting multiple ``expert models'' whose outputs are fused in a single prediction, as 
in the work by Blum~\etalcite{blum2018iros}.

\section{Approach to \mbox{RGB-D} Panoptic Segmentation}
\label{sec:main}

Our panoptic segmentation network is an encoder-decoder architecture that operates on \mbox{RGB-D} images and processes RGB and depth data 
by means of two different encoders. Encoders features are merged at different output strides, 
and are sent to three 
decoders that restore the backbone features to the original image
resolution. The first decoder targets semantic segmentation. The second decoder
predicts the location of object centers in the form of a probability heatmap. The third decoder 
predicts an embedding vector for each pixel of the image. Finally, a post-processing
module aggregates information coming from the last two decoders to 
obtain instance segmentation in a bottom-up fashion. \figref{fig:architecture} illustrates our proposed network architecture. 
The next sections explain the individual parts of our method.

\subsection{Encoders}
Our panoptic segmentation network is based on two ResNet34 encoders~\cite{he2016cvpr}, which are fed 
with the RGB image $I_{\mathrm{rgb}} \in \mathbb{R}^{3 \times H \times W}$ and the depth image
$I_{\mathrm{depth}} \in \mathbb{R}^{1 \times H \times W}$, respectively.  
In both encoders, the basic ResNet block is replaced by the Non-Bottleneck-1D block~\cite{romera2018tits}, which allows a more lightweight architecture 
than the vanilla ResNet, since all $3 \times 3$ convolutions are replaced by 
a sequence of $3 \times 1$ and $1 \times 3$ convolutions with a ReLU in between, while increasing segmentation
performance~\cite{seichter2021icra}. We merge features from the
two encoders at different output strides and project them into the RGB encoder.
We provide more details about our merging strategy in Sec.~\ref{sec:feature_fusion}. After the last 
merging, the resulting feature is processed by an adaptive pyramid
pooling module~\cite{zhao2017cvpr-pspn}, which has the role of increasing 
the receptive field of the network. From the RGB encoder, we extract features at different output strides 
and use them in the decoders by means of skip connections~\cite{ronneberger2015micc}.

\subsection{Feature Fusion}
\label{sec:feature_fusion}
We perform feature fusion in the encoders at different output strides. 
We merge features 
from the two encoders at every downsampling step, and then send them to the 
RGB encoder. The depth encoder processes
depth features only, to avoid processing the same features with both encoders.

\begin{figure}[t]
  \centering
  \includegraphics[width=0.99\linewidth]{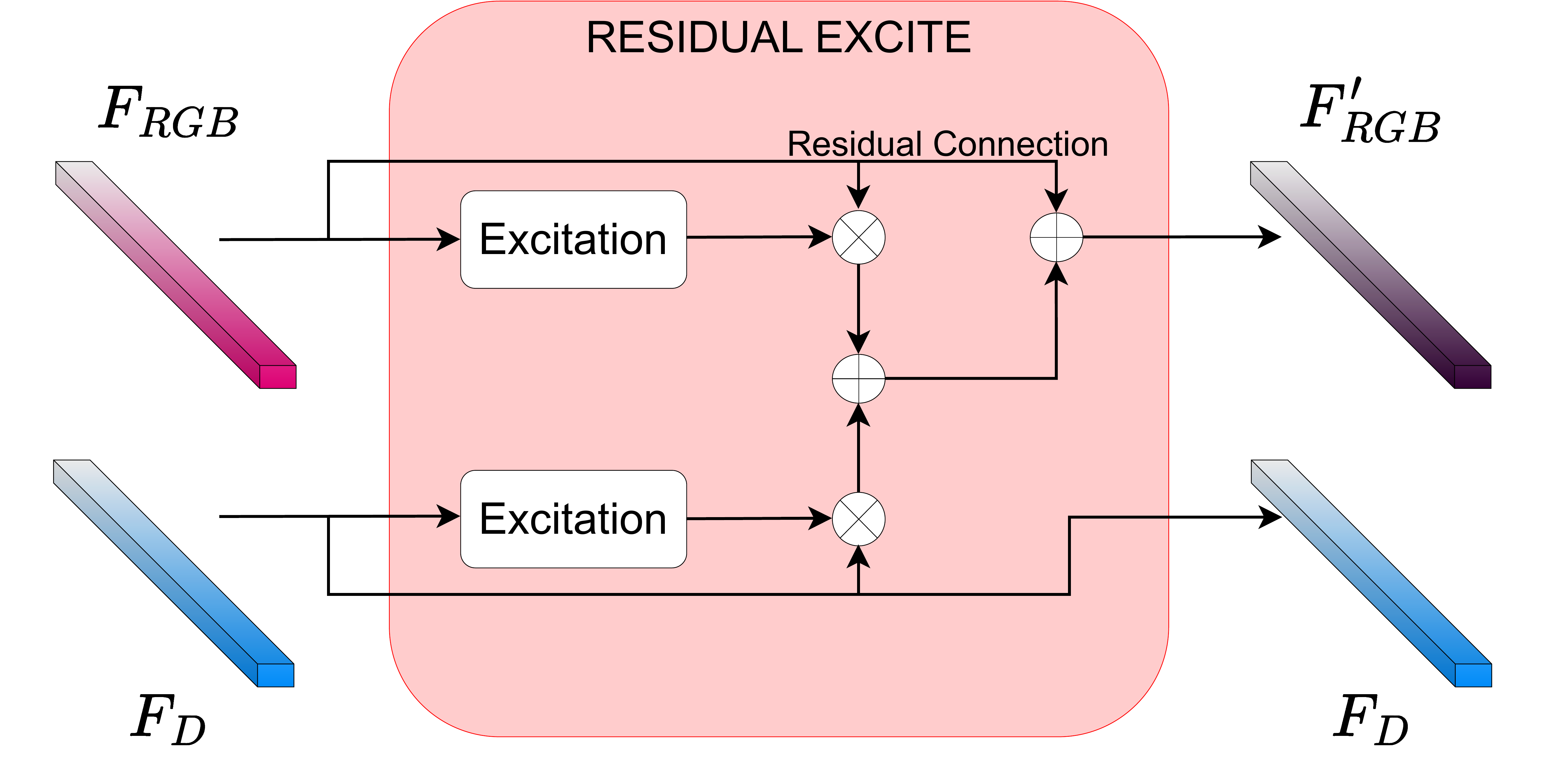}
  \caption{Detail of the ResidualExcite module. It elaborates the
  feature maps and produces a novel one that encodes information from both RGB and depth. Symbols $\bigoplus$ and $\bigotimes$ stand for elementwise addition and multiplication, respectively.}
  \label{fig:fusion_modules}
\end{figure}

We propose a novel way of merging 
features, inspired by the Squeeze-and-Excitation module~\cite{hu2018cvpr-sn}.
This module produces a channel descriptor (squeezing operation), and assigns 
to each channel a modulation weight that is finally applied to the feature map (excitation).
Our goal is to obtain a global modulation weight rather than a channelwise weight, as we believe that a more fine-grained reweighing of features is crucial for effective segmentation results. Thus, we remove the squeezing operation, and
we add a residual connection. This module, called ResidualExcite (see~\figref{fig:fusion_modules}), is given by
\begin{equation}
  \b{\mathrm{X}}_{\mathrm{rgb}} \hspace{-0.1em} = \hspace{-0.1em} \b{\mathrm{X}}_{\mathrm{rgb}} + \lambda \, \big(E(\b{\mathrm{X}}_{\mathrm{rgb}}) \, \b{\mathrm{X}}_{\mathrm{rgb} }
  + E(\b{\mathrm{X}}_\mathrm{depth}) \, \b{\mathrm{X}}_{\mathrm{depth}} \big),
\end{equation}
where $\b{\mathrm{X}}_i \in \mathbb{R}^{C_d \times H_d \times W_d}, \, i \in \{ \mathrm{rgb}, \mathrm{depth} \}$ is the feature 
coming from the respective branch, 
$E(\b{\mathrm{X}}_i) \in \mathbb{R}^{C_d \times H_d \times W_d}$ is the excitation module, which is a sequence of $1 \times 1$ convolutions followed
by a sigmoid activation function, $\lambda$ is a (non-trained) parameter for weighing the excitation module over 
the residual connection, and the subscript $d$ refers to the dimension of the features at the specific 
output stride in which the merging happens. The RGB and the depth features are both 
individually excited (meaning both excitation and elementwise multiplication) 
and then summed, so that each of them can be used separately in case the other cue is missing. 
Finally, a residual connection adds $\b{\mathrm{X}}_{\mathrm{rgb}}$ again.

\subsection{Decoders}
The decoders are composed of three SwiftNet-like modules~\cite{orsic2019cvpr}, where we incorporate Non-Bottleneck-1D blocks, and we extend the feature channel to $512$ in the first module and then we reduce it as the resolution increases. Finally, two upsampling modules based on nearest-neighbor and depthwise convolutions, that are less computationally expensive than transposed convolutions~\cite{seichter2021icra}, restore the original resolution.
Our model is composed of three decoders, for semantic segmentation, center prediction, and embedding prediction.

\textbf{Semantic Segmentation}. The semantic segmentation decoder has an output depth 
equal to the number of semantic classes $C$, $I_{\mathrm{sem}} \in \mathbb{R}^{C \times H \times W}$, 
and a softmax activation function.
It is trained with the usual cross-entropy loss $\mathcal{L}_{\mathrm{sem}}$ for one-hot encoded multi-label classification.

\textbf{Center Prediction}. The center prediction decoder has an output depth of $1$, 
$I_{\mathrm{cen}} \in \mathbb{R}^{1 \times H \times W}$, and a sigmoid activation
function to predict pixelwise probabilities of being a center. It is optimized with a binary focal loss~\cite{milioto2019icra-fiass}:
\begin{equation}
  \mathcal{L}_{\mathrm{cen}} = \begin{cases}
    - \alpha \, (1 - \hat{y})^\tau \, \log(\hat{y}) \qquad \qquad , \,\mathrm{if} \ y = 1,\\
    - (1 - \alpha) \, \hat{y}^\tau \, \log(1 - \hat{y}) \qquad \, \, , \,\mathrm{otherwise},
  \end{cases}
  \label{eq:binaryfocalloss}
\end{equation}
where $\alpha$ and $\tau$ are design parameters and are fixed in all experiments to 
$0.1$ and $2$, respectively.

\textbf{Embedding Prediction}. The third decoder of the network predicts a $D_{\mathrm{emb}}$-dimensional 
embedding vector \mbox{$I_{\mathrm{emb}} \in \mathbb{R}^{D_{\mathrm{emb}} \times H \times W}$} for each pixel in
the image, and is optimized with a composed hinged loss. The first term $\mathcal{L}_{\mathrm{att}}$
attracts embedding vectors of pixels belonging to the same instance, the second 
term $\mathcal{L}_{\mathrm{rep}}$ repel embedding vectors of pixels belonging to different instances, and the third
term $\mathcal{L}_{\mathrm{reg}}$ is a regularization term that avoids exploding entries:
\begin{align}
  \label{eq:composedhingedloss}
  \mathcal{L}_{\mathrm{emb}} & = \beta_1 \, \mathcal{L}_{\mathrm{att}} + 
                                  \beta_2 \, \mathcal{L}_{\mathrm{rep}} +
                                  \beta_3 \, \mathcal{L}_{\mathrm{reg}},\\
  \mathcal{L}_{\mathrm{att}} & = \dfrac{1}{K} \sum_{k=1}^K \, \dfrac{1}{P_k} \, \sum_{p=1}^{P_k} \, \big[\norm{\hat{\b{e}}_k - \hat{\b{e}}_p} - \delta_{\mathrm{a}} \big]^+, \\
  \mathcal{L}_{\mathrm{rep}} & = \dfrac{1}{K(K-1)} \underset{k_1 \neq k_2}{\sum_{k_1=1}^K \, \sum_{k_2=1}^{K-1}} \, \big[\delta_{\mathrm{r}} - \norm{\hat{\b{e}}_{k_1} - \hat{\b{e}}_{k_2}} \big]^+,
\end{align}
  \begin{align}
\mathcal{L}_{\mathrm{reg}} & = \dfrac{1}{K} \, \sum_{k=1}^K \, \norm{\hat{\b{e}}_k},
\end{align}
where $\hat{\b{e}}_i \in \mathbb{R}^{D_{\mathrm{emb}}}$ is the unbounded logit predicted by the decoder, $K$ is the number 
of instances in the image, $P_k$ is the number of pixels of the specific 
instance, $[\cdot]^+$ corresponds to $\max(0, \cdot)$, and $\delta_{\mathrm{a}}$ and $\delta_{\mathrm{r}}$ are thresholds for 
attracting and repelling the embedding vectors, respectively. To speed up computations, we compute
$\mathcal{L}_{\mathrm{att}}$ only between pixels belonging to an instance and their corresponding center,
and $\mathcal{L}_{\mathrm{rep}}$ only among centers of different instances. Similarly, we regularize
only the vectors of the centers. 

We optimize the network with a panoptic loss that is a weighted sum of the previously-defined terms:
\begin{equation}
  \mathcal{L}_{\mathrm{pan}} = w_{\mathrm{1}} \, \mathcal{L}_{\mathrm{sem}} +
                               w_{\mathrm{2}} \, \mathcal{L}_{\mathrm{cen}} +
                               w_{\mathrm{3}} \, \mathcal{L}_{\mathrm{emb}}.
\end{equation}

\subsection{Post-processing}
Our post-processing 
module computes the instance mask based on the output of the three decoders. Since the center prediction
decoder usually outputs blobs around the desired center, we perform 
a non-maximum suppression operation in order to reduce each blob to a single pixel, filtered by the semantic prediction to ensure consistency. 

In particular, centers are first filtered
by the semantic prediction $I_{\mathrm{sem}}$ to avoid having centers belonging to stuff classes, 
which
do not have any instance. Then, pixels that have a probability of being a center lower than
a threshold~$\delta_{\mathrm{cen}}$ are discarded. 
Next, for each blob, we extract the pixel with the highest probability of being a center.
A blob $\mathcal{B}$ is defined as the set of pixels belonging to the
same semantic class and having a similar embedding vector. Referring to $\Omega$ as the set of pixels who are predicted as 
centers in $I_{\mathrm{cen}}$, i.e., $\Omega = \{p \mid I_{\mathrm{cen}}(p) \geq \delta_{\mathrm{cen}}\}$, a blob is defined as 
\begin{equation}
  \begin{split}
    \mathcal{B} = \big\{ p_i, \, p_j \in \Omega  &\mid I_{\mathrm{sem}}(p_i) = c
    \wedge I_{\mathrm{sem}}(p_j) = c\\
    & \quad \wedge \, \norm{\hat{\b{e}}_{p_i} - \hat{\b{e}}_{p_j}} < \delta_{\mathrm{emb}}, \ i \neq j\big\},
  \end{split}
\end{equation}
where $c$ is a specific semantic class, $\delta_{\mathrm{emb}}$ is a threshold for aggregating embedding vectors, and $p_i, \, p_j$ are generic pixels.

After the center extraction, we perform an agglomerative clustering 
operation to group pixels to centers according to the Euclidean distance in the embedding space 
and semantic class. For each center, we compute its distance in the embedding space from all pixels of the same semantic class.
This operation is less computationally intensive than the similarity matrix between all pixels of the image, and motivates the use of object centers.
Finally, we assign the pixel to a center if 
their distance in the embedding space is below a threshold~$\theta$. The use of the semantic segmentation prediction enforces
consistency and avoids grouping pixels belonging to different semantic classes in the same object.

\begin{figure*}[ht!]
  \centering
  \includegraphics[width=0.99\linewidth]{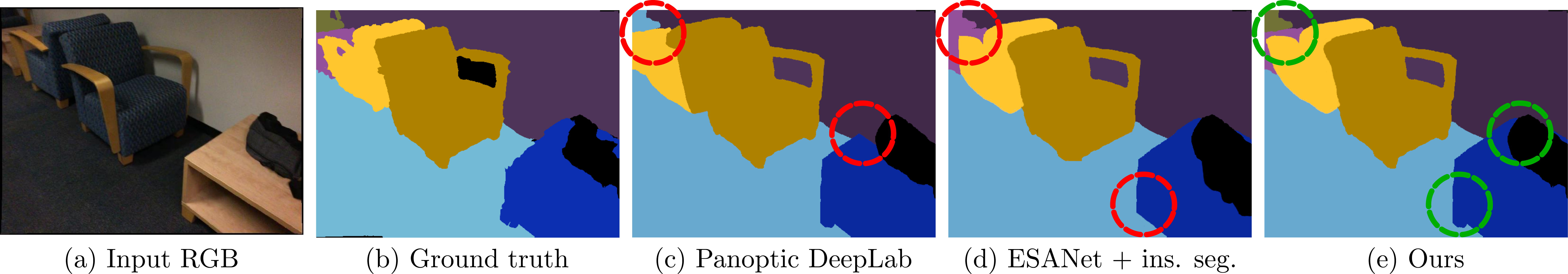}
  \caption{Experimental results on ScanNet.
  Our approach achieves superior segmentation results when compared 
  to the baselines.}
  \label{fig:experiments1}
\end{figure*} 
\begin{figure*}[ht]
  \centering
  \includegraphics[width=0.99\linewidth]{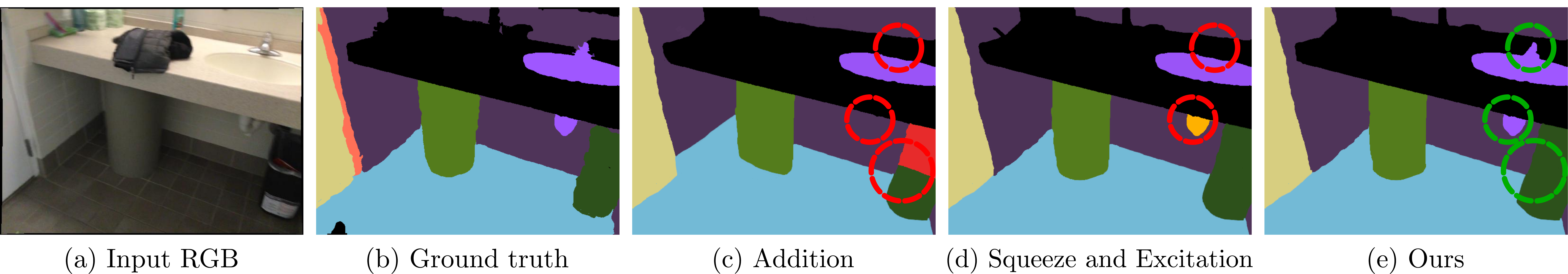}
  \caption{Experimental results on ScanNet. 
  Our approach achieves superior segmentation results when compared 
  to other fusion modules.}
  \label{fig:experiments2}
  \vspace{-0.5em}
\end{figure*}

\subsection{Robustness to Missing Inputs}
\label{sec:input_invariance}
Since we process RGB and depth with two separate encoders, it is possible to 
feed the network with partial information, i.e., without either RGB or depth, and freeze the part 
corresponding to the missing data. This can be done also at training
time, with a switching mechanism that freezes gradients if no input is provided to 
one branch. In this way, the frozen encoder does not contribute to the 
forward and backward pass, and the network can train at the same time with 
complete \mbox{RGB-D}, RGB-only, or depth-only images. Furthermore, the network is able to infer on 
different data without the need for re-training. 
Feature merging with partial data is not an issue, since the remaining cue 
can still be excited (or squeezed and excited) and processed. 

We train the full model with a probability of dropping data (RGB or depth),
equal to $p_{\mathrm{drop}}$. This means that 
the network can train either with the full RGB-D data or not. If data is dropped, then no 
input is sent to the corresponding encoder, which we freeze. Additionally,
we use an adaptive sampling mechanism to choose what needs to be dropped: in particular, if one cue has been dropped 
more times than the other, its probability of being dropped in the next iteration is reduced. This helps having a more 
balanced dropping mechanism and alleviates the problem of dropping always the same modality.

\section{Experimental Evaluation}
\label{sec:exp}

%

%
We present our experiments to show the capabilities of our method and compare it with other fusion methods common in the literature.
Furthermore, we show performance of models trained with partial data.

\subsection{Experimental Setup}

\textbf{Datasets and Metrics}.
We test our method on the validation sets of two datasets: ScanNet~\cite{dai2017cvpr} and HyperSim~\cite{roberts2021iccv-haps}. 
ScanNet is composed of $~$2.5M real-world images 
organized in 1,513 scenes.
HyperSim is a photorealistic synthetic dataset of indoor scenes, and it is composed of 
77.4K images organised in 461 scenes. 
For both datasets, we do not consider stuff classes (wall, floor) for instance segmentation.

For the center prediction, we 
pre-process the instance masks of both datasets to extract a center ground truth that is inside the object mask. We consider this to be more effective than 
computing the center of the associated bounding box, which can fall outside the object mask and the segmentation mask, 
for example in the case of an isolated concave object.

We evaluate our method by means of the panoptic quality~(PQ)~\cite{kirillov2019cvpr} and the mean intersection over union 
(mIoU)~\cite{everingham2010ijcv} over all classes for semantic segmentation.

\begin{figure*}[t]
  \centering
  \includegraphics[width=0.99\linewidth]{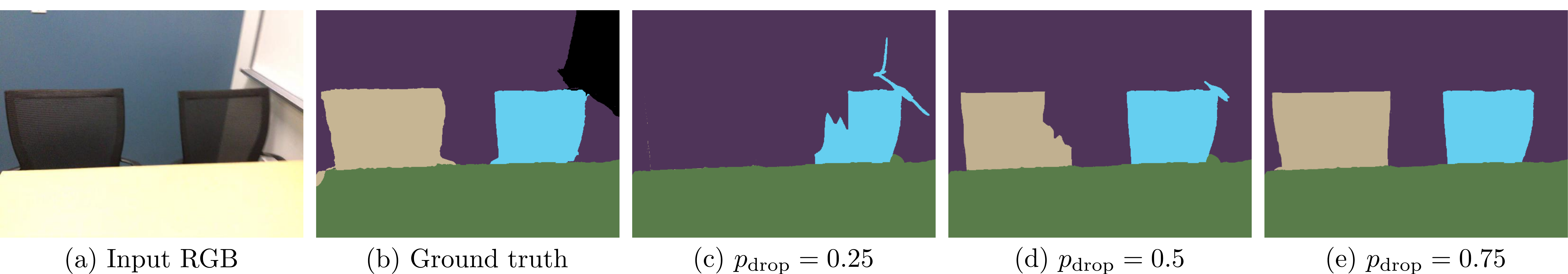}
  \caption{Results when doing inference on RGB-only after training with missing inputs. The bigger $p_{\mathrm{drop}}$, the 
  better the performance.}
  \label{fig:experiments3}
\end{figure*}
\textbf{Training details and parameters}.
In all experiments, except when explicitly specified, we use the one-cycle learning rate 
policy~\cite{smith2019aiml} with an initial learning rate of 0.004. We perform random scale, crop, and flip data 
augmentation, and optimize with AdamW~\cite{loshchilov2017arxiv}, for 200 epochs. The batch size is set to 32.
Additionally, we set $D_{\mathrm{emb}} = 32$ as embedding dimension, $\delta_a = 0.1$, $\delta_r = 1$, 
$\delta_{\mathrm{emb}} = 0.5$, $\delta_{\mathrm{cen}} = 0.5$, $\theta = 0.5$, and 
$\lambda = 1.5$. Loss weights are set to 
$w_1 = 1$, $w_2 = 0.1$, $w_3 = 10$, $\beta_1 = 1$, $\beta_2 = 1$, $\beta_3 = 0.001$.
 



\begin{table}[t]
  \centering
  
  \begin{tabular}{C{4cm}ccc}
    \toprule
    Method                                              & Dataset            & PQ                  & mIoU    \\
    \toprule
    RGB Panoptic DeepLab~\cite{cheng2020cvpr-pass}      & ScanNet            & 30.11                & 43.12 \\
    RGB-D Panoptic DeepLab                              & ScanNet            & 31.43                & 45.45 \\
    ESANet~\cite{seichter2021icra} with Addition        & ScanNet            & 35.65                & 51.78 \\
    ESANet~\cite{seichter2021icra} with SE~\cite{hu2018cvpr-sn}            & ScanNet            & 37.09                & 54.01 \\
    Ours with CBAM~\cite{woo2018eccv}                   & ScanNet            & 39.11                & 58.11 \\
    Ours with ResidualExcite                            & ScanNet            & \b{40.87}          & \b{58.98} \\
    \midrule
    RGB Panoptic DeepLab~\cite{cheng2020cvpr-pass}      & HyperSim           & 26.10                & 40.45 \\
    RGB-D Panoptic DeepLab                              & HyperSim           & 28.56                & 41.08 \\
    ESANet~\cite{seichter2021icra} with Addition        & HyperSim           & 32.18                & 50.74 \\
    ESANet~\cite{seichter2021icra} with SE~\cite{hu2018cvpr-sn}            & HyperSim           & 35.87                & 54.07 \\
    Ours with CBAM~\cite{woo2018eccv}                   & HyperSim           & 37.02                & 54.21 \\
    Ours with ResidualExcite                            & HyperSim           & \b{38.67}            & \b{55.14} \\
    \bottomrule
  \end{tabular}
  \caption{Performance of the different panoptic segmentation methods. Best result
  in bold.}
  \label{tab:baselines}
\end{table}

\subsection{Panoptic Segmentation on RGB-D Images}
The first set of experiments evaluates the performance of our proposed method, and offers comparisons to 
other architectures common in the literature. 
We base our work on ESANet~\cite{seichter2021icra}, which is a double-encoder 
network for \mbox{RGB-D} semantic segmentation on images. 
To use it as a baseline for panoptic segmentation, we expand ESANet with the decoders for the center prediction and 
embedding prediction. Notice that ESANet leverages Squeeze-and-Excitation as a fusion strategy, but reports in the paper also fusion by addition 
that simply sums up features coming from the two encoders and 
projects them into the RGB encoder. Here, we use both variants. Additionally, we use another fusion module as baseline, CBAM~\cite{woo2018eccv}, which infers attention maps 
along two separate dimensions, channel, and spatial.
Furthermore, we also compare against a single-encoder architecture that process the RGB-D image as a four-channel input signal. For that, we
adapted Panoptic DeepLab~\cite{cheng2020cvpr-pass} to process images with four channels, and we fed the model with a 4D tensor that is the 
concatenation of the RGB and the depth images. 

We compare 
our approach to such methods since we focus on image-like data, without relying on 3D representations such as truncated signed distance fields or point clouds.
Results are reported in \tabref{tab:baselines}, qualitative results are shown in \figref{fig:experiments1} and \figref{fig:experiments2}. 
Our reimplementation of Panoptic DeepLab shows inferior performance when 
compared to ESANet and our approach. We also report numbers from the vanilla implementation of Panoptic DeepLab (called RGB Panoptic DeepLab in the Table)
that does not make use of the depth. Interestingly, the performance of the RGB Panoptic DeepLab is close to the one of the
RGB-D re-implementation, that simply processes an input with four channels rather than three. This suggests that
processing depth as an additional input channel does not add much information, while a separate processing
via a second encoder is more effective for such a task.
The ResidualExcite module helps segmentation performance, and outperforms other merging strategies such as CBAM
and Squeeze-and-Excitation (ESANet + instance segmentation). Fusion by addition shows inferior performance, which is an expected result as it
processes all features without weighing them according to their effective usefulness. This experiment indicates that our
more fine-grained weighing mechanism, which has effect on each single entry of the encoder feature rather than
each channel, enhances performance of the downstream task. Additionally, our network provides end-to-end predictions at 10 Hz, that need to be post-processed to obtain the final instance segmentation mask.

To empirically validate our architecture design, we compare it with some state-of-the-art models from the ScanNet benchmark
for semantic segmentation~\cite{hazirbas2016accv, valada2019ijcv}. We use both our full model and its task-specific reduction, in which the decoders for center 
and embedding prediction are cut out in order to do semantic segmentation only. Table~\ref{tab:semantic} shows that even if our full model has 
weaker performance, our task specific model outperforms the baselines. Note that some methods higher in the benchmark rely on multiple 
frames as input, and thus cannot be directly compared to our approach.

\begin{table}[t]
  \centering
  
  \begin{tabular}{C{3.4cm}ccc}
    \toprule
    Method                  & Dataset           & mIoU \\
    \toprule
    AdapNet++~\cite{valada2019ijcv}             & ScanNet           & 54.61 \\
    FuseNet~\cite{hazirbas2016accv}             & ScanNet           & 56.65 \\
    SSMA~\cite{valada2019ijcv}                  & ScanNet           & 66.13 \\
    Ours (full model)       & ScanNet           & 58.98 \\
    Ours (semantic)         & ScanNet           & \b{69.78} \\
    \bottomrule
  \end{tabular}
  \caption{Performance of different semantic segmentation models on the ScanNet dataset. 
  Best result in bold.}
  \label{tab:semantic}
  \vspace{-0.8em}
\end{table}

  

  \subsection{Experiments on Robustness to Missing Inputs}
  \label{sec:ivd}
  The second set of experiments backs up our claim that our approach can train and infer on
  partial data, such that the network learns to deal with missing RGB- or depth-frames. 
  We test different values for $p_{\mathrm{drop}}$: 0.25, 0.5, and 0.75. This means that 
  the network will drop either the RGB frame or the depth frame according to the specified probability.
  If dropping happens, we choose which cue to drop according to the adaptive sampling mechanism 
  mentioned in Sec.~\ref{sec:input_invariance}. This strategy gives a better performance than random sampling,
  which is therefore not reported here. 
  \tabref{tab:input_invariance} shows performance for inference on \mbox{RGB-D}, RGB-only, and depth-only data. All models 
  produce inferior segmentation results than the model that does not drop any frame (same model discussed above), when doing inference on 
  full \mbox{RGB-D} frames. However, its performance drops substantially when doing inference on partial data,
  as the network was never trained with missing cues. 
  Additionally, we notice how dropping frames more often makes the model better for doing inference on partial data, since the network 
  trained more with missing cues. On the contrary, low values of $p_{\mathrm{drop}}$ bring poor performance when handling 
  partial data, because the network was mainly trained with both, RGB and depth. The model trained with $p_{\mathrm{drop}} = 0.5$
  is the best compromise to achieve satisfactory results both on RGB-D, RGB-only, and depth-only, even without reaching 
  the performance of the RGB-D model. Qualitative results are shown 
  in \figref{fig:experiments3}.
  
  \begin{table}[t]
    \centering
    \begin{tabular}{ccccccc}
      \toprule
                               & \multicolumn{2}{c}{RGB-D}             & \multicolumn{2}{c}{RGB-only}   & \multicolumn{2}{c}{Depth-only}\\
      \cmidrule(lr){2-3} \cmidrule(lr){4-5} \cmidrule(lr){6-7}
      \textit{p}\textsubscript{drop}      & PQ                  & mIoU            & PQ                  & mIoU     & PQ       & mIoU\rule{0pt}{3ex} \\
      \toprule
      0                        & \b{40.87}           & \b{58.98}       & 9.12                & 21.13    & 12.54    & 24.57 \\
      0.25                     & 31.12               & 44.55           & 20.12               & 34.83    & 21.18    & 35.14   \\
      0.5                      & 30.73               & 42.86           & 25.61               & 39.18    & 26.74    & 38.87  \\
      0.75                     & 26.81               & 40.07           & \b{27.48}           & \b{39.56}& \b{28.18}& \b{39.45} \\
      \bottomrule
    \end{tabular}
    
    \caption{Performance of the model when dropping either RGB or depth with different probability. Best result in bold.}
    \label{tab:input_invariance}
    \vspace{-0.3cm}
  \end{table}
All experiments described in Sec.~\ref{sec:ivd} are done with a batch size of 4 and an initial learning rate of 0.001. 
Due to the missing inputs, the training procedure is less stable and thus required a smaller learning rate. We use ResidualExcite for merging; 
experiments are performed on ScanNet only.

\subsection{Ablation Studies}
In this last section, we provide ablations to show the improvements provided by the fusion strategy.
We perform all ablations on the ScanNet dataset only.

First, we analyze the 
ResidualExcite and investigate the effect of the residual connection. Without it, the excitation module (ExciteOnly)
still provides an entrywise reweighing of the feature. Experiments show 
that this is already enough to ensure superior performance with respect to other 
baselines, but the residual connection gives further improvements, see \tabref{tab:ablations}.
Additionally, in our case, fusing in the RGB encoder is more effective than fusing in 
the depth encoder.

\begin{table}[t]
  \centering
  
  \begin{tabular}{C{0.62cm}C{0.62cm}C{0.62cm}C{0.62cm}C{0.62cm}|cc}
    \toprule
    RGB           & D           & SE              & E             & RE              & PQ          & mIoU        \\
    \toprule
    \checkmark    &             &                 &               &                 & 25.63        & 38.91 \\
                  & \checkmark  &                 &               &                 & 28.89        & 41.01 \\
    \textcircled{$\checkmark$}    & \checkmark  &                 &               &                 & 35.65        & 51.78 \\
    \textcircled{$\checkmark$}    & \checkmark  & \checkmark      &               &                 & 37.09        & 54.01 \\
    \textcircled{$\checkmark$}    & \checkmark  &                 & \checkmark    &                 & 38.73        & 55.57 \\
    \checkmark    &\textcircled{$\checkmark$}     &                 &               & \checkmark  & 38.80    & 56.67 \\        
    \textcircled{$\checkmark$}    & \checkmark  &                 &               & \checkmark      & \b{40.87}    & \b{58.98} \\        
    \bottomrule
  \end{tabular}
  
  \caption{The first two lines refer to RGB- and depth-only. Then, we show double-encoder networks with 
  addition (RGB + D), Squeeze-and-Excitation (SE), ExciteOnly (E) and ResidualExcite (RE).
  We use \textcircled{$\checkmark$} to indicate which branch processes fused features.}
  \label{tab:ablations}
\end{table}
In the same table, we compare the performance of the full model reductions in which a single encoder is used. 
We test panoptic segmentation on RGB-only and depth-only data. The results are clearly inferior to the double-encoder models. 
Interestingly, depth-only gives better results than RGB-only. This is probably due to the fact that some scenes have 
challenging lighting conditions, and some objects are hard to recognize in the RGB image. Such 
information is not lost in the depth image. Also, this suggests that geometric cues may be more relevant than color information 
when it comes to object recognition for segmentation.



\section{Conclusion}
\label{sec:conclusion}
In this paper, we presented a novel approach for panoptic segmentation on \mbox{RGB-D} images based on a 
double encoder architecture with intermediate feature merging. 
Our method exploits the inner structure of the neural network to enable training and inference when cues are missing 
using the same model and without the need for retraining.
We implemented and evaluated our approach on different datasets
and provided comparisons with other existing models and supported
all claims made in this paper. The experiments suggest that our more fine-grained reweighing 
of features is crucial for effective segmentation results. Additionally, models trained with partial data 
achieve inferior performances on \mbox{RGB-D} segmentation when compared with full models, but they work better when 
inferring on partial data.


\section*{Acknowledgments}
We thank Andres Milioto and Xieyuanli Chen for their constructive feedback and useful discussions. 

\bibliographystyle{plain_abbrv}

\bibliography{glorified,new}

\end{document}